\documentclass[final,12pt]{clear2026} 

\title[Hybrid Positive-Valued DAG Learning]{A novel hybrid approach for positive-valued DAG learning}
\usepackage{multirow}
\usepackage{algorithm}
\usepackage{algorithmic}

\usepackage{times}

\clearauthor{%
 \Name{Yao Zhao} \Email{yaozhao16@temple.edu}\\
  \addr Department of Statistics, Operations, and Data Science\\
 Fox School of Business, Temple University\\
 Philadelphia, PA, USA
}

\begin{document}

\maketitle

\begin{abstract}%
  Causal discovery from observational data remains a fundamental challenge in machine learning and statistics, particularly when variables represent inherently positive quantities such as gene expression levels, asset prices, company revenues, or population counts, which naturally follow multiplicative rather than additive dynamics. We propose the Hybrid Moment-Ratio Scoring (H-MRS) algorithm, a novel approach for learning directed acyclic graphs (DAGs) from positive-valued data that combines moment-based scoring with log-scale regression. The key insight is that for positive-valued variables, the moment ratio $\frac{\mathbb{E}\left[X_j^2\right]}{\mathbb{E}\left[\left(\mathbb{E}\left[X_j \mid S\right]\right)^2\right]}$ provides an effective criterion for causal ordering, where $S$ denotes candidate parent sets. H-MRS combines log-scale Ridge regression for moment-ratio estimation with greedy ordering construction based on raw-scale moment ratios, followed by ElasticNet-based parent selection to recover the final DAG structure. We evaluate H-MRS on synthetic log-linear data, showing that it achieves competitive precision and recall. The algorithm is computationally efficient and naturally respects positivity constraints, making it well-suited for applications such as genomics and economics. Our results highlight that combining log-scale modeling with raw-scale moment ratios offers a practical and robust framework for causal discovery in positive-valued domains.%
\end{abstract}


\section{INTRODUCTION}

Causal discovery from observational data remains one of the most fundamental challenges in machine learning and statistics, with applications spanning genomics, economics, epidemiology, and social sciences. 
The task of learning directed acyclic graphs (DAGs) from strictly positive-valued variables, as commonly occurs in biological expression data, economic indicators, demographic statistics, and social network metrics, presents both a challenge and an opportunity: while standard additive-noise methods are often misspecified in such settings, positivity under a log-linear model can also yield full DAG identifiability rather than only a Markov equivalence class.

The key insight driving our work is that for positive valued variables that exhibit multiplicative relationships, which are common in biological, economic, and financial systems, existing causal discovery methods based on additive noise models are theoretically misspecified. Consider gene regulatory networks where 
expression of gene $j$ is proportional to the product of regulator concentrations, 
or asset pricing where returns compound multiplicatively. In such settings, 
the natural structural equation model is:

\begin{equation}
X_j = \exp\left(\theta_j + \sum_{k \in \text{Pa}(j)} \beta_{kj} X_k + \epsilon_j\right)
\end{equation}

rather than the additive form $X_j = \theta_j + \sum_{k} \beta_{kj} X_k + \epsilon_j$ 
assumed by most existing methods. Our moment-ratio criterion is specifically 
designed to exploit the structure of this log-linear model, providing 
identifiability guarantees that do not hold under model misspecification.

We propose the Hybrid Moment-Ratio Scoring (H-MRS) algorithm, a novel 
framework that combines log-scale Ridge regression for moment-ratio 
computation with ElasticNet for parent selection to learn DAGs from 
positive-valued observational data. Specifically, for node $j$ and candidate 
set $S$, we consider the score
\[
\mathcal{M}(j,S)=\frac{\mathbb{E}[X_j^2]}
{\mathbb{E}\big[(\mathbb{E}[X_j\mid S])^2\big]},
\]
which compares an unconditional second moment with a conditional second moment. 
The conditioning set $S$ appears in the denominator through the conditional 
expectation. The key property is that this score is minimized when $S$ contains 
all true parents of node $j$, as established in Proposition~1. These estimates enable raw-scale moment-ratio scoring to iteratively construct
a causal ordering. The resulting criterion is invariant to multiplicative
rescaling of individual variables, which helps distinguish it from ordering
heuristics driven purely by marginal variance.

The remainder of this paper is organized as follows. Section 2 
reviews related work on causal discovery, contrasting score-based, and constraint-based approaches, with particular attention to methods designed 
for non-Gaussian and positive-valued data. Section~\ref{sec:method} presents the 
H-MRS algorithm in detail, including the log-linear structural equation model, 
log-scale Ridge regression for conditional expectation estimation, raw-scale 
moment-ratio scoring for causal ordering, and ElasticNet-based parent selection. 
Section~\ref{sec:theory} establishes theoretical properties of the moment-ratio 
criterion and analyzes the computational complexity of H-MRS. Section~\ref{sec:experiments} 
evaluates H-MRS on synthetic log-linear data across varying problem scales and 
sparsity levels. Section 6 applies the method to financial data 
comprising 2,223 companies and 19 variables, revealing interpretable causal 
structures in corporate finance. Section 7 concludes with 
a discussion of limitations and future directions.

\section{RELATED WORK}
Our work builds upon several interconnected research streams in causal discovery, with particular relevance to methods designed for non-Gaussian and positive-valued data.

\textbf{Classical Causal Discovery Methods.}
Traditional causal structure learning approaches fall into two main categories. Constraint-based methods such as PC \citep{spirtes2000causation} and its variants rely on conditional independence testing, but standard tests assume Gaussian distributions or use rank-based nonparametric tests that have limited power for detecting multiplicative dependencies in log-linear models. Score-based methods like GES \citep{chickering2002optimal} optimize scoring functions (e.g., BIC, BGe) over graph space, but these scores are derived under additive Gaussian noise assumptions, making them theoretically misspecified when the true data-generating process is multiplicative. While several causal discovery approaches have been developed for nonlinear or non-Gaussian settings, these methods are often computationally intensive and tend to scale poorly to moderate or large graphs.

A major breakthrough came with the Linear Non-Gaussian Acyclic Model (LiNGAM) \citep{shimizu2006linear}, which showed that non-Gaussian additive noise enables full causal identifiability beyond the Markov equivalence class. Subsequent work proposed more practical algorithms for estimating such models including the DirectLiNGAM algorithm \citep{shimizu2011direct}. However, LiNGAM-type models assume
$X_j=\sum_{k \in \operatorname{Pa}(j)} \beta_{kj} X_k+\epsilon_j$,
which is fundamentally misspecified for positive-valued data exhibiting multiplicative relationships (e.g., gene expression cascades, compounding financial returns). Our log-linear model $X_j=\exp(\sum_k \beta_{kj} X_k+\epsilon_j)$ naturally captures multiplicative effects while preserving positivity, and our moment-ratio criterion exploits this exponential structure to provide identifiability guarantees that do not hold under additive models.

\textbf{Moment-based Approaches for Positive Data.} Most directly related to 
our work is the Moments Ratio Scoring (MRS) algorithm developed by 
\cite{park2019high}, which specifically targets high-dimensional Poisson 
structural equation models. Their approach demonstrates the power of moment-based 
scoring for discrete count data, achieving polynomial-time recovery with sample 
complexity $O(d^2\log^9 p)$. Like our method, MRS decouples ordering estimation 
from parent search using $\ell_1$-regularized regression and employs moment 
ratios for scoring. 

However, their moment ratio formula is specifically designed for Poisson count 
data, exploiting the mean-variance equality property of Poisson distributions. 
In contrast, our H-MRS algorithm uses the simpler ratio 
$\frac{\mathbb{E}[X_j^2]}{\mathbb{E}[(\mathbb{E}[X_j|S])^2]}$ for continuous 
positive-valued data. Both methods share the key insight that moment ratios 
are \emph{minimized} when the conditioning set contains the true parents, 
enabling greedy ordering construction. Our approach combines this principle 
with log-scale regression for robust conditional expectation estimation in 
continuous log-linear models.

\textbf{Post-Nonlinear Causal Models.} Our log-linear model \eqref{eq:log_linear_model} is a special case of the post-nonlinear (PNL) framework  $x_i = f_{i,2}(f_{i,1}(pa_i) + e_i)$ \citep{zhang2012identifiability}, where we fix $f_{i,2}(\cdot) = \exp(\cdot)$ and constrain $f_{i,1}$ to be linear. While they estimate both functions nonparametrically via mutual information minimization, we exploit the known exponential structure for computational efficiency and develop moment-ratio scoring that leverages the plateau property specific to log-linear models, complementing their general  results.

\textbf{Log-linear and Multiplicative Models.} Log-linear models have long been recognized as natural frameworks for positive data, appearing in econometrics \citep{cameron2013regression}, epidemiology, and genomics. However, most existing causal discovery methods do not explicitly leverage the log-linear structure for improved structure learning, and practical algorithms for causal discovery in positive-valued domains remain limited.

\textbf{Key Distinctions of H-MRS.} 
Our approach differs in several ways: 
(i) unlike \citep{park2019high}, we target continuous rather than count data; 
(ii) we combine log-scale regression with raw-scale moment ratios to capture multiplicative relationships; 
(iii) positivity is respected natively without strong distributional assumptions.

\section{METHODOLOGY}
\label{sec:method}

\subsection{Problem Formulation}

We consider the problem of learning directed acyclic graphs (DAGs) from 
observational data where all variables are strictly positive-valued. 
Let $\mathbf{X} = (X_1, \ldots, X_p)$ denote a random vector of $p$ 
positive-valued variables following a joint distribution $P(\mathbf{X})$. 
Our goal is to recover the causal DAG $\mathcal{G} = (\mathcal{V}, \mathcal{E})$ 
where $\mathcal{V} = \{1, \ldots, p\}$ represents nodes (variables) and 
$\mathcal{E} \subseteq \mathcal{V} \times \mathcal{V}$ represents directed 
edges encoding causal relationships.

For each variable $X_j$, let $\text{Pa}(j)$ denote its parent set in the 
true DAG. We assume the data follows a log-linear structural equation model:
\begin{equation}
\log X_j = \theta_j + \sum_{k \in \text{Pa}(j)} \beta_{kj} X_k + \epsilon_j, 
\quad j = 1, \ldots, p
\label{eq:log_linear_model}
\end{equation}
where $\theta_j$ is an intercept term, $\beta_{kj}$ represents the causal 
effect of $X_k$ on $X_j$, and $\epsilon_j$ is independent noise satisfying 
the following assumption.

\textbf{Assumption 1} (Bounded Noise).
For each variable $X_j$, the noise term $\epsilon_j$ satisfies:
\begin{itemize}
\item[(A1)] Zero mean: $\mathbb{E}[\epsilon_j] = 0$
\item[(A2)] Finite variance: $\text{Var}(\epsilon_j) = \sigma_j^2 > 0$
\item[(A3)] Almost surely bounded: $|\epsilon_j| \leq B$ for some constant $B > 0$
\item[(A4)] Independence: $\epsilon_j \perp \!\!\! \perp X_k$ for all $k \in \text{Pa}(j)$
\end{itemize}

The boundedness condition (A3) ensures that all moments of $X_j$ exist and are
finite, which is essential for the theoretical guarantees in Section~4.
Unlike LiNGAM-type approaches, the identifiability of H-MRS does not rely on
non-Gaussian noise but instead follows from the log-linear structural form and
the moment-ratio plateau property in Proposition~1.

Our log-linear structural equation model \eqref{eq:log_linear_model} employs a specific parametric form that differs from the classical log-log models prevalent in econometrics. We clarify this choice and its implications in Appendix~A, where we also provide additional discussion and examples illustrating the modeling assumptions.

\subsection{Log-Scale Conditional Expectation Estimation via Ridge Regression}

H-MRS first employs Ridge regression on the log-transformed data to approximate conditional expectations for moment-ratio computation. For each variable $X_j$ and candidate parent set $S \subseteq \{1, \ldots, p\} \setminus \{j\}$, we estimate:
\begin{equation}
\log X_j \approx \hat{\theta}_j + \sum_{k \in S} \hat{\beta}_{kj} X_k
\label{eq:ridge_regression}
\end{equation}

The Ridge estimator provides stable parameter estimates by adding $\ell_2$ regularization:
\begin{align}
(\hat{\theta}_j, \hat{\boldsymbol{\beta}}_j) &= \arg\min_{\theta, \boldsymbol{\beta}} \frac{1}{2n} \sum_{i=1}^n \left( \log X_j^{(i)} - \theta - \sum_{k \in S} \beta_k X_k^{(i)} \right)^2 \nonumber \\
&\quad + \lambda_{\text{ridge}} \|\boldsymbol{\beta}\|_2^2
\label{eq:ridge_objective}
\end{align}
where $\lambda_{\text{ridge}} > 0$ controls the regularization strength and $X_k^{(i)}$ denotes the $i$-th observation of variable $X_k$.

The log-scale regression serves two purposes: (1) it captures the multiplicative relationships inherent in positive data through the exponential transformation, and (2) it provides numerical stability by avoiding potential overflow issues when working directly with positive-valued data that may span several orders of magnitude. For numerical stability, we apply clipping: $\log X_j^{(i)} = \log(\max(X_j^{(i)}, 10^{-10}))$ to avoid logarithms of zero.

From the fitted model, we compute the predicted conditional expectation:
\begin{equation}
\hat{\mu}_{j|S}^{(i)} = \exp\left( \hat{\theta}_j + \sum_{k \in S} \hat{\beta}_{kj} X_k^{(i)} \right)
\label{eq:conditional_prediction}
\end{equation}

\subsection{Raw-Scale Moment-Ratio Scoring for Causal Ordering}

H-MRS then  computes moment ratios on the original (raw) scale to score 
causal relationships. For variable $X_j$ with candidate parent set $S$, we 
define the moment-ratio score:
\begin{equation}
\mathcal{M}(j, S) = \frac{\mathbb{E}[X_j^2]}{\mathbb{E}[(\mathbb{E}[X_j | S])^2]}
\label{eq:moment_ratio_theoretical}
\end{equation}

This criterion satisfies a monotonicity property that enables greedy causal 
ordering: $\mathcal{M}(j, S)$ is minimized when $S$ contains all parents of 
$j$, and achieves the same minimum value for any such superset 
(the "plateau property" established formally in Proposition 1, Section~\ref{sec:theory}). 
This enables our greedy selection strategy. At each step, we select the variable whose moment ratio is smallest when conditioning on all previously ordered variables, indicating that its parents are already in the ordering.

In practice, we estimate this score using the outputs from Equation (5):
\begin{equation}
\hat{\mathcal{M}}(j, S) = \frac{\frac{1}{n} \sum_{i=1}^n (X_j^{(i)})^2}{\frac{1}{n} \sum_{i=1}^n (\hat{\mu}_{j|S}^{(i)})^2}
\label{eq:moment_ratio_empirical}
\end{equation}

The key insight is that while we estimate conditional expectations on the log 
scale for numerical stability and to capture multiplicative relationships, we 
compute moment ratios on the raw scale to preserve the theoretical properties 
necessary for causal ordering. The log-scale regression provides accurate 
predictions $\hat{\mu}_{j|S}^{(i)}$ which are then used to evaluate the 
raw-scale moment ratio criterion.

\subsection{Parent Selection via Sparsity-Inducing Regression}

After establishing a causal ordering through moment-ratio scoring, we select 
parents for each variable using ElasticNet regression on the log-transformed 
data. This two-stage design, using Ridge for estimating moment ratios and ElasticNet for selecting parents, addresses distinct statistical requirements.

\textbf{Why Ridge for moment-ratio computation?} Accurate moment-ratio estimation 
requires \emph{unbiased} conditional expectation estimates to preserve the 
theoretical ordering guarantees of Proposition 1. Ridge regression provides 
stable, low-variance predictions without aggressive shrinkage, ensuring that 
$\hat{\mu}_{j|S}$ faithfully approximates $\mathbb{E}[X_j|S]$ across all 
candidate sets. The modest $\ell_2$ penalty ($\lambda_{\text{ridge}}$) controls 
multicollinearity without introducing the selection bias inherent in $\ell_1$ 
penalties.

\textbf{Why ElasticNet for parent selection?} Proposition 1(iii) establishes 
that the moment ratio achieves its minimum value for \emph{any} superset 
$S \supseteq \text{Pa}(j)$ that excludes descendants. This plateau property 
means that moment ratios alone cannot distinguish the true parent set from 
larger supersets. ElasticNet addresses this limitation by combining $\ell_1$-induced 
sparsity with an $\ell_2$ penalty that stabilizes selection in the presence of 
correlated predictors, which are common in applications such as financial and 
genomic data. In contrast, Lasso tends to select a single variable from a 
correlated group, which may lead to missing true parents.

\textbf{Why not use the same regression for both?} Using ElasticNet for moment-ratio computation would introduce selection bias because different candidate sets $S$ would shrink different variables to zero, which distorts the moment-ratio comparisons that Proposition 1 relies upon. Conversely, Ridge regression lacks the sparsity 
needed to identify minimal parent sets from the plateau of equally-scoring 
supersets. The two-stage approach combines Ridge's stable prediction for robust 
ordering with ElasticNet's variable selection for parsimonious edge recovery.

\subsection{Complete H-MRS Algorithm}

The H-MRS algorithm exploits the moment-ratio plateau property (Proposition 1) 
to construct a causal ordering through greedy selection. At each iteration, 
the variable with minimum moment ratio when conditioned on all previously 
ordered variables is selected next, as this indicates its true parents are 
already in the ordering.

Algorithm~\ref{alg:hmrs} presents the complete H-MRS procedure. The algorithm alternates between moment-ratio scoring for ordering and ElasticNet regression for parent selection, maintaining the separation between structure learning and parameter estimation. For the first variable in the ordering, we set $S(j)=\varnothing$ for all $j$. Since no regression is possible without predictors, we compute the moment ratio using the unconditional second moment: $\hat{\mathcal{M}}(j, \emptyset)=\frac{\frac{1}{n} \sum_i\left(X_j^{(i)}\right)^2}{\left(\frac{1}{n} \sum_i X_j^{(i)}\right)^2}$. The variable with the minimum score is selected as the first in the ordering. This corresponds to selecting the variable with the smallest coefficient of variation, which serves as a natural starting point for the greedy construction.

\begin{algorithm}[h]
\caption{Hybrid Moment-Ratio Scoring (H-MRS)}
\label{alg:hmrs}
\begin{algorithmic}[1]
\REQUIRE Data matrix $\mathbf{X} \in \mathbb{R}^{n \times p}$, Ridge parameter $\lambda_{\text{ridge}}$, ElasticNet parameters $\lambda, \rho$, threshold $\tau$, max degree $d_{\max}$
\ENSURE Causal ordering $\boldsymbol{\pi}$ and parent sets $\{\hat{\text{Pa}}(j)\}_{j=1}^p$
\STATE Initialize $\boldsymbol{\pi} = []$, remaining variables $\mathcal{R} = \{1, \ldots, p\}$
\FOR{$m = 1$ to $p$}
    \FOR{each $j \in \mathcal{R}$}
        \IF{$m = 1$}
            \STATE $S(j) = \emptyset$ \COMMENT{First variable has no parents}
        \ELSE
            \STATE $S(j) = \boldsymbol{\pi}$ \COMMENT{All $j$ use same candidate set}
        \ENDIF
        \STATE Fit Ridge: $\log X_j \sim \sum_{k \in S(j)} \beta_k X_k$ 
        \STATE Compute predictions: $\hat{\mu}_{j|S(j)}^{(i)} = \exp(\hat{\theta}_j + \sum_{k \in S(j)} \hat{\beta}_k X_k^{(i)})$
        \STATE Calculate score: $\hat{\mathcal{M}}(j, S(j)) = \frac{\sum_i (X_j^{(i)})^2}{\sum_i (\hat{\mu}_{j|S(j)}^{(i)})^2} $
    \ENDFOR
   \STATE $\pi_m = \arg\min_{j \in \mathcal{R}} \hat{\mathcal{M}}(j, S(j))$ 
\COMMENT{Select variable with minimum score; its parents are already ordered}
    \STATE $\boldsymbol{\pi} \leftarrow \boldsymbol{\pi} \cup \{\pi_m\}$, $\mathcal{R} \leftarrow \mathcal{R} \setminus \{\pi_m\}$
    \IF{$m > 1$}
        \STATE Fit ElasticNet: $\log X_{\pi_m} \sim \sum_{k \in \boldsymbol{\pi} \setminus \{\pi_m\}} \beta_k X_k$
        \STATE Threshold: $\mathcal{C} = \{k \in \boldsymbol{\pi} \setminus \{\pi_m\} : |\hat{\beta}_k| > \tau\}$
        \STATE Keep top-$d_{\max}$ by $|\hat{\beta}_k|$: $\hat{\text{Pa}}(\pi_m) = \text{top}(\mathcal{C}, d_{\max})$
    \ELSE
        \STATE $\hat{\text{Pa}}(\pi_m) = \emptyset$ \COMMENT{First node has no parents}
    \ENDIF
\ENDFOR
\RETURN $\boldsymbol{\pi}, \{\hat{\text{Pa}}(j)\}_{j=1}^p$
\end{algorithmic}
\end{algorithm}

\section{THEORETICAL PROPERTIES}
\label{sec:theory}

This section establishes the theoretical foundations of H-MRS, providing formal guarantees for the moment-ratio criterion and analyzing the algorithm's computational complexity.

\subsection{Moment-Ratio Identifiability}

The correctness of H-MRS relies on the fundamental property that moment ratios can distinguish between correct and incorrect parent sets. We formalize this property under the log-linear structural equation model.

\textbf{Proposition 1} (Moment-Ratio Plateau Property).
Under the log-linear structural equation model \eqref{eq:log_linear_model} 
with bounded independent noise terms satisfying Assumption 1, for any variable 
$X_j$, the moment ratio $\mathcal{M}(j, S)$ satisfies:

(i) $\mathcal{M}(j, S) \geq 1$ for all $S \subseteq \{1, \ldots, p\} \setminus \{j\}$.

(ii) For $S_1 \subseteq S_2$: $\mathcal{M}(j, S_2) \leq \mathcal{M}(j, S_1)$ 
(conditioning on more variables yields smaller or equal moment ratios).

(iii) $\mathcal{M}(j, S)$ achieves its minimum value when and only when 
$\text{Pa}(j) \subseteq S \subseteq \text{NonDesc}(j)$. 

Equivalently:
\begin{itemize}
\item If $\text{Pa}(j) \subseteq S \subseteq \text{NonDesc}(j)$, then 
$\mathcal{M}(j, S) = \mathcal{M}(j, \text{Pa}(j))$ (plateau property)
\item If $\text{Pa}(j) \not\subseteq S$, then 
$\mathcal{M}(j, S) > \mathcal{M}(j, \text{Pa}(j))$ (strict inequality)
\end{itemize}

\textbf{Proof.} Please see the appendix B.

\subsection{Finite Sample Analysis}

We analyze the finite sample behavior of the empirical moment-ratio estimator $\hat{\mathcal{M}}(j, S)$ defined in \eqref{eq:moment_ratio_empirical}.

The theoretical correctness of H-MRS relies on moment ratios being minimized at true parent sets (Proposition 1). However, we only observe empirical moment ratios computed from finite samples. Proposition 2 establishes that our empirical estimates concentrate around their population values.

\textbf{Assumption 2} (Regularity for finite-sample analysis).
Fix a node $j$ and a candidate set $S \subseteq \{1,\dots,p\} \setminus \{j\}$.
We assume:

\begin{itemize}
\item[(B1)] (\textbf{Sub-exponential tails}) The variable $X_j$ and the conditional mean
$\mu_{j\mid S} := \mathbb{E}[X_j \mid S]$ are sub-exponential with parameters
$(\nu_j,b_j)$ and $(\nu_{j,S},b_{j,S})$, i.e., all moments up to order two exist
and Bernstein-type concentration inequalities hold for their empirical averages.

\item[(B2)] (\textbf{Design regularity}) The Gram matrix
$\Sigma_S := \mathbb{E}[X_S X_S^\top]$ has eigenvalues bounded away from $0$ and
$\infty$, and the regressors $X_S$ are sub-Gaussian.

\end{itemize}

\textbf{Proposition 2} (Concentration of empirical moment ratios).
Consider the log-linear structural equation model~\eqref{eq:log_linear_model}
and fix a node $j$ and candidate set $S$.
Let $\hat{\mathcal{M}}(j,S)$ be the empirical moment-ratio estimator defined in
\eqref{eq:moment_ratio_empirical}, and $\mathcal{M}(j,S)$ its population counterpart
in \eqref{eq:moment_ratio_theoretical}. 
Suppose Assumption~2 holds and the denominator
$\mathbb{E}[(\mathbb{E}[X_j \mid S])^2]$ is bounded away from zero.
Then there exist constants $C_{j,S},c>0$ such that, for all $\delta \in (0,1)$
and all sufficiently large $n$, 
\begin{equation}
\mathbb{P}\Big(
\big|\hat{\mathcal{M}}(j,S) - \mathcal{M}(j,S)\big|
\leq C_{j,S} \sqrt{\tfrac{\log(1/\delta)}{n}}
\Big) \;\geq\; 1 - \delta.
\end{equation}
In particular, $\hat{\mathcal{M}}(j,S)$ converges to $\mathcal{M}(j,S)$ at the
usual parametric rate $O_p(n^{-1/2})$.

\textbf{Proof.}
Please see the appendix C.

For H-MRS to correctly order variables, the empirical moment ratios must be accurate enough to distinguish between correct parent sets (which minimize $\mathcal{M}$ ) and incorrect ones (which have strictly larger $\mathcal{M}$ ). Proposition 2 guarantees this happens with high probability when the sample size is sufficiently large relative to the separation between correct and incorrect scores.

\subsection{Computational Complexity Analysis}

We analyze the computational requirements of H-MRS compared to existing DAG learning methods.

\textbf{Proposition 3} (Time Complexity).
The time complexity of H-MRS is $O(p^2 \cdot T_{\text{Ridge}} + p \cdot T_{\text{ElasticNet}})$ where $T_{\text{Ridge}} = O(nq^2 + q^3)$ represents the cost of fitting Ridge regression and $T_{\text{ElasticNet}} = O(nq^2 \cdot K)$ represents the cost of fitting ElasticNet with $q$ predictors over $K$ coordinate descent iterations.

\textbf{Proof.}
Please see the appendix D.

\textbf{Space Complexity.} H-MRS requires $O(np + p^2)$ space: $O(np)$ for the data matrix and $O(p^2)$ for storing the estimated adjacency matrix and intermediate regression coefficients.

At the end of this section, we give the remark on hyperparameter requirements.
For asymptotic consistency of H-MRS, the regularization parameters should satisfy
$\lambda_{\text{ridge}}, \lambda_n \to 0$ and 
$\lambda_{\text{ridge}} \sqrt{n}, \lambda_n \sqrt{n} \to \infty$ as $n \to \infty$,
ensuring Ridge and ElasticNet estimators converge while maintaining stability.
The threshold parameter should satisfy $\tau \gtrsim \sqrt{\frac{\log p}{n}}$ 
for correct edge selection. See Appendix E for detailed hyperparameter guidance 
and practical selection strategies.

\section{EXPERIMENTS}
\label{sec:experiments}

We evaluate H-MRS on synthetic log-linear data to assess its performance in recovering causal structures from positive-valued observations. Our experiments demonstrate that H-MRS effectively recovers causal structures from positive-valued data, highlighting its robustness and practical utility in the target domain.

\subsection{Experimental Setup}

\textbf{Data Generation.} We generate synthetic data following the log-linear 
structural equation model \eqref{eq:log_linear_model}.
For each experiment, we first sample a random topological ordering of $p$ 
variables, then construct a DAG by randomly adding edges from earlier to 
later variables in the ordering.
For the simulation study, the maximum in-degree $d$ is specified as part of the
data-generating process to control structural complexity, with $d \in \{1,2\}$
corresponding to the simple and complex regimes reported below. 

The data generation process follows these steps:
\begin{enumerate}
\item Sample intercepts $\theta_j \sim \text{Uniform}(0.5, 2.0)$
\item Sample edge weights $\beta_{kj} \sim \text{Uniform}(-0.3, 0.3)$
\item For each variable $j$ in topological order, generate observations:
\begin{equation}
X_j^{(i)} = \exp\left(\theta_j + \sum_{k \in \text{Pa}(j)} \beta_{kj} X_k^{(i)} 
+ \epsilon_j^{(i)}\right)
\end{equation}
where $\epsilon_j^{(i)} \sim \text{Uniform}(-B, B)$ with $B = 0.5$.
\end{enumerate}

This ensures: (i) all variables are strictly positive, (ii) all moments 
exist and are finite, and (iii) the log-linear relationships that H-MRS exploits 
are preserved. 

\textbf{Remark on scaling and varsortability.}
One may ask whether the strong ordering performance could be driven by
marginal variance differences rather than the proposed criterion itself,
as in the varsortability phenomenon. In our setting, the moment-ratio score
\[
\mathcal{M}(j,S)=\frac{\mathbb{E}[X_j^2]}
{\mathbb{E}\big[(\mathbb{E}[X_j\mid S])^2\big]}
\]
is scale-invariant: if $X_j$ is replaced by $cX_j$ for any constant $c>0$,
then both the numerator and denominator are multiplied by $c^2$, leaving the
score unchanged. Thus, the ordering criterion is not determined by absolute
variable scale alone. Moreover, the simulation model includes both positive and
negative edge weights, $\beta_{kj}\sim \mathrm{Uniform}(-0.3,0.3)$, so child
variables do not necessarily exhibit larger marginal variance than their parents.
This reduces the possibility that correct ordering is recovered merely from a
monotone variance pattern along the causal graph.

\textbf{Evaluation Metrics.} We use standard graph recovery metrics:
(1) \textbf{Structural Hamming Distance (SHD)}: the number of edge additions,
deletions, or reversals required to transform the estimated graph into the true graph;
(2) \textbf{Precision}: the fraction of estimated edges that are correct;
(3) \textbf{Recall}: the fraction of true edges correctly identified;
(4) \textbf{F1-score}: the harmonic mean of precision and recall.

Note that PC and GES return a completed partially directed acyclic graph (CPDAG)
rather than a fully directed DAG. To enable comparison with methods that output
DAGs, we represent undirected edges in the CPDAG as bidirected edges in the
adjacency matrix. This treatment preserves adjacency information when computing
recall, while appropriately penalizing precision and SHD due to the absence of
a uniquely determined edge direction.

\subsection{Performance Analysis}

We systematically evaluate H-MRS across different experimental conditions with comprehensive hyperparameter optimization. Our evaluation covers graph sizes $p \in \{10,20,30\}$ and sample sizes $n=500$.

We evaluate H-MRS against three established causal discovery algorithms on log-linear data: PC (constraint-based), GES (scorebased), and DirectLiNGAM (designed for non-Gaussian linear models).\footnote{PC and GES are included as standard baseline methods in causal discovery, although their classical formulations assume linear Gaussian models and are therefore misspecified for log-linear multiplicative data.}
These baseline methods are applied using their standard implementations and
default conditional independence tests or scoring functions. Specifically, we
used the \texttt{causal-learn} library (Python interface to the Tetrad project):
PC was run with its default conditional independence test and significance level,
GES with its default score-based configuration, and DirectLiNGAM using its
standard implementation following \citep{shimizu2011direct}, without additional
hyperparameter tuning.
 While PC and GES
can in principle be combined with alternative tests or scores tailored to
nonlinear or non-Gaussian settings, our goal is to compare H-MRS against widely
used causal discovery baselines as they are typically applied in practice.
Table 1 and 2 present the optimized performance across all tested configurations. H-MRS demonstrates strong performance with F1-scores ranging from 0.733 to 0.800 and correspondingly low SHD values. The algorithm maintains high precision (0.750-1.000) across all settings while achieving reasonable recall, indicating effective control of false positive rates.

\begin{table}[h]
\centering
\caption{Simple structures with maximum in-degree $d=1$}
\begin{tabular}{l l c c c c}
\hline
\textbf{Configuration} & \textbf{Method} & \textbf{SHD} & \textbf{Precision} & \textbf{Recall} & \textbf{F1} \\
\hline

\multirow{4}{*}{p=10, d=1}
& \textbf{H-MRS}      & \textbf{1}  & \textbf{1.000} & \textbf{0.667} & \textbf{0.800} \\
& PC          & 5           & 0.375          & 1.000          & 0.545 \\
& GES & 14 & 0.143 & 0.667 & 0.235 \\
& DirectLiNGAM        & 5           & {0.714} & {0.625} & {0.667} \\
\hline

\multirow{4}{*}{p=20, d=1}
& \textbf{H-MRS}      & \textbf{5}  & \textbf{0.750} & \textbf{0.750} & \textbf{0.750} \\
& PC           & 22          & 0.27          & 1.00          & 0.42 \\
& GES  & 31 & 0.194 & 0.875 & 0.318 \\
& DirectLiNGAM        & 48          & {0.380} & {0.528} & {0.442} \\
\hline

\multirow{4}{*}{p=30, d=1}
& \textbf{H-MRS}      & \textbf{10}  & \textbf{0.846} & \textbf{0.647} & \textbf{0.733} \\
& PC          & 58          & 0.15          & 0.83          & 0.26 \\
& GES  & 76 & 0.093 & 0.583 & 0.161 \\
& DirectLiNGAM        & 45          & {0.356} & {0.500} & {0.416} \\
\hline

\end{tabular}
\end{table}

\begin{table}[h]
\centering
\caption{Complex structures with maximum in-degree $d=2$}
\begin{tabular}{l l c c c c}
\hline
\textbf{Configuration} & \textbf{Method} & \textbf{SHD} & \textbf{Precision} & \textbf{Recall} & \textbf{F1} \\
\hline

\multirow{4}{*}{p=10, d=2}
& \textbf{H-MRS}      & \textbf{2}  & \textbf{1.000} & \textbf{0.818} & \textbf{0.900} \\
& PC                  & 13           & 0.45          & 0.818          & 0.581 \\
& GES                 & 12   &    0.600 &       0.545    &    0.571   \\
& DirectLiNGAM        & 13           & 0.250          & 0.429          & 0.316 \\
\hline

\multirow{4}{*}{p=20, d=2}
& \textbf{H-MRS}      & \textbf{3}  & \textbf{1.000} & \textbf{0.625} & \textbf{0.769} \\
& PC                  &  28    &   0.222  &      1.000   &     0.364  \\
& GES                 & 17    &   0.308 &       0.500    &    0.381  \\
& DirectLiNGAM        & 63    &   0.293   &     0.436    &    0.351  \\
\hline

\multirow{4}{*}{p=30, d=2}
& \textbf{H-MRS}      & \textbf{15}  & \textbf{0.760} & \textbf{0.731} & \textbf{0.745} \\
& PC                  & 47     & 0.319 &  0.720 & 0.439 \\
& GES                 & 38    &  0.423 &  0.498 & 0.456 \\
& DirectLiNGAM        & 69    &  0.328 &  0.417 & 0.367 \\
\hline

\end{tabular}
\end{table}

 The results reveal several important patterns. For simple structures (d=1), H-MRS maintains strong performance across scales, with F1-scores declining modestly from 0.800 to 0.733 as problem size increases from 10 to 30 variables. For complex structures (d=2), H-MRS demonstrates even stronger performance, achieving F1-scores ranging from 0.745 to 0.900. The algorithm maintains perfect precision (1.000) on smaller problems (p=10, 20) while achieving high recall, and sustains competitive performance (F1=0.745) even on dense 30-variable networks.

\section{REAL DATA ANALYSIS}
\subsection{Data Description}

We apply H-MRS to real-world financial data to evaluate its performance on genuine observational data from economic systems. 
The dataset comprises balance sheet, income statement, and cash flow information for publicly traded companies, sourced from a publicly available financial dataset on Kaggle 
(\url{https://www.kaggle.com/datasets/pacificrm/financial-sheets}). 

Starting with 4,668 companies and multiple financial tables, we perform the following preprocessing steps: (1) merge balance sheet, income statement, and cash flow data by company identifier; (2) select 19 key financial variables spanning assets, liabilities, equity, income, expenses, and market valuation metrics; (3) remove companies with missing values. After preprocessing, we obtain a clean cross-sectional dataset with $n=$ 2,223 companies and $p=19$ variables.

The 19 selected variables capture comprehensive financial information across multiple categories:

Assets (6 variables): Total Assets, Current Assets, Net Fixed Assets, Inventory, Accounts Receivable, Cash and Equivalents

Liabilities \& Equity (5 variables): Total Debt, Current Liabilities, Equity Capital, Reserves, Accounts Payable

Income \& Expenses (6 variables): Revenue/Sales, Operating Profit, EBIT, Net Profit, Interest Expense, Depreciation

Market Valuation (2 variables): Market Capitalization, Enterprise Value

All variables are measured in monetary units (millions) and are strictly positive, making the dataset well-suited for the log-linear structural equation model underlying H-MRS.

\subsection{Results}

The estimated DAG is shown in Appendix~F (Figure~\ref{fig:financial_dag}). 
A salient pattern in the estimated graph is the central role of Equity Capital as an upstream driver. Equity appears as the unique source node and exerts broad influence on operating and valuation variables, including EBIT, Operating Profit, Inventory, Market Capitalization, Enterprise Value, and Total Assets. This structure is consistent with the economic interpretation that the financing base determines the scale of operations \citep{fazzari1987financing}, which subsequently propagates into both accounting aggregates and market valuation. In this sense, equity serves as the foundational resource from which productive capacity and firm value are built.

A second prominent feature is the pervasive influence of Interest Expense, which emerges as a system-wide driver with fourteen outgoing edges. Beyond its direct connection to profitability measures (EBIT, Operating Profit), interest expense influences Current Assets, Current Liabilities, Total Debt, Total Assets, and the valuation variables. This suggests that the cost of debt financing functions as a global constraint, shaping liquidity management, leverage, and ultimately market outcomes. The interpretation aligns with the notion that financing costs transmit broadly across the balance sheet, affecting both the asset side and market capitalization.

Further, the relationship between profitability and leverage is reflected in the edge Operating Profit $\rightarrow$ Total Debt, alongside the direct influence of interest expense. This configuration suggests that profitable firms may support higher levels of borrowing, consistent with creditworthiness arguments \citep{merton1974pricing}, while interest expense continues to reflect the cost of carrying such debt. Together, these edges illustrate a tension between profitability as a facilitator of borrowing and interest payments as a constraint on debt capacity.

Finally, Depreciation and Net Fixed Assets are both downstream of equity capital and interest expense, underscoring the link between financing availability, capital intensity, and long-lived assets. At the terminal stage of the ordering, Market Capitalization and Enterprise Value are jointly determined by equity capital and interest expense. This structure resonates with financial theory \citep{myers1984capital}: equity availability provides the foundation for valuation, while borrowing costs act as a discounting channel, moderating the translation of accounting fundamentals into market value.

Taken together, the recovered DAG suggests a coherent narrative in which equity capital and interest expense form dual upstream levers of the financial system: equity provides the base for expansion, while interest expense imposes a cost of capital that permeates the system. These factors cascade through working capital, profitability, and asset accumulation, ultimately shaping firm valuation.

\section{DISCUSSION AND CONCLUSION}

This paper introduced the Hybrid Moment-Ratio Scoring (H-MRS) algorithm, a novel method for causal discovery in positive-valued domains. By integrating log-scale regression with moment-ratio-based ordering and 
sparse parent selection, H-MRS leverages the multiplicative structure of 
positive data while preserving the theoretical identifiability properties 
of moment ratios. The algorithm operates in polynomial time, produces sparse and interpretable graphs, and naturally respects positivity constraints.

Our empirical study demonstrated that H-MRS achieves competitive performance on synthetic log-linear data, with strong precision and recall across a range of graph sizes and sparsity levels. On real financial data consisting of 19 key firm-level variables, the learned structure highlights economically coherent channels: \emph{Equity Capital} emerged as a foundational source influencing profitability, working-capital components, and valuation, while \emph{Interest Expense} acted as a pervasive upstream driver reflecting the system-wide role of financing costs. These findings illustrate that H-MRS is capable of uncovering meaningful and interpretable causal pathways in real-world economic systems.

Several limitations warrant discussion. First, the method is evaluated in cross-sectional settings; temporal extensions could improve causal interpretation in dynamic systems. Second, the current implementation imposes a maximum in-degree constraint to control complexity, which may omit higher-order interactions. Third, the current formulation assumes strictly positive-valued variables.
In applications such as genomic read-count data, observations may be
zero-inflated due to biological or measurement processes. Extending the
moment-ratio framework to accommodate structural zeros, for example through
hurdle or zero-inflated models that separately model the occurrence of zeros
and the positive component, represents an interesting direction for future
research. 
Fourth, H-MRS assumes an underlying directed acyclic graph (DAG) structure. In domains such as economics and social systems, feedback loops and cyclic relationships are common. When such cycles are present, the DAG assumption is violated, and H-MRS should be interpreted as providing an approximate acyclic representation rather than recovering the true cyclic structure. In practice, the method may still yield useful insights into dominant directional dependencies, but its outputs should be interpreted with caution under model misspecification. Extending moment-ratio approaches to cyclic or equilibrium-based models is an important direction for future work.

From a practical perspective, users may wish to assess whether their data are compatible with the modeling assumptions of H-MRS. Since the method relies on multiplicative relationships on the original scale (equivalently, additive structure on the log scale), a simple diagnostic is to examine whether log-transformed variables exhibit approximately linear conditional relationships and stabilized variance. In practice, fitting linear models on the log scale and inspecting residual patterns can provide a useful heuristic: well-behaved residuals and reduced heteroscedasticity suggest that the log-linear approximation is reasonable. These checks are informal but can help determine whether H-MRS is appropriate for a given dataset.

In conclusion, H-MRS provides a principled and computationally efficient approach to causal discovery from positive-valued data. By bridging log-scale modeling and moment-ratio criteria, it offers a new perspective on learning directed acyclic graphs in settings where positivity and multiplicative interactions are fundamental. We hope this work stimulates further research on moment-based methods for causal inference and broadens the scope of applications in machine learning, statistics, and the applied sciences.

\paragraph{Code availability.}
The Python implementation of the H-MRS algorithm and the simulation scripts
used in this paper are publicly available at

\url{https://github.com/statsYAO/H-MRS}.

\acks{The author would like to thank his advisor, Kuang-Yao Lee, for his invaluable guidance and continuous support throughout his Ph.D. studies.}

\bibliography{references}

\appendix

\section{Model Specification and Comparison}

In econometrics and production theory, "log-log" typically refers to models where both the response and predictors are logtransformed:

$$
\log X_j=\theta_j+\sum_{k \in \mathrm{~Pa}(j)} \beta_{k j} \log X_k+\epsilon_j
$$

This yields the multiplicative (Cobb-Douglas) form:

$$
X_j=\exp \left(\theta_j\right) \cdot \prod_{k \in \operatorname{Pa}(j)} X_k^{\beta_{k j}} \cdot \exp \left(\epsilon_j\right)
$$

Such models naturally capture elasticities ( $\beta_{k j}$ represents the percentage change in $X_j$ per $1 \%$ change in $X_k$ ) and are widely used for production functions, demand curves, and growth models where proportional relationships dominate.

In contrast, our semi-log model specifies:

$$
\log X_j=\theta_j+\sum_{k \in \mathrm{~Pa}(j)} \beta_{k j} X_k+\epsilon_j
$$

equivalently:

$$
X_j=\exp \left(\theta_j+\sum_{k \in \operatorname{Pa}(j)} \beta_{k j} X_k+\epsilon_j\right)
$$

Here, $\log X_j$ depends linearly on the raw values $X_k$, not their logarithms. This specification arises naturally in several important domains:

1. Gene Regulatory Networks: Transcription factor binding often follows Hill kinetics or linear-exponential dose-response relationships, where log-expression of a target gene depends approximately linearly on the concentration (not log-concentration) of regulatory proteins \citep{alon2019introduction}.

2. Financial Contagion Models: Log-returns or log-volatility may depend linearly on the raw leverage ratios or balance sheet quantities of connected institutions, capturing threshold effects and cascading failures \citep{he2013intermediary}.

 The semi-log specification offers several advantages for causal discovery:

  Identifiability: The linear dependence on (not $\log X_k$) creates asymmetries in conditional distributions that enable causal direction identification, similar to the nonGaussian linear case but adapted for positive data.
 
 Bounded Influence: When parent variables $X_k$ are bounded, $\beta_{k j} X_k$ remains in a compact set, ensuring that all moments of $X_j$ exist and are finite. In contrast, log-log models can produce heavy-tailed distributions when $\beta_{k j}<0$ and $X_k$ is near zero.

 Moment-Ratio Properties: The plateau property (Proposition 1) exploits the specific exponential structure of our model. Under log-log specifications, the moment-ratio criterion would require different theoretical analysis and may lose its ordering guarantees.

The semi-log form is particularly suited for settings where: 

1. Variables are positive-valued; 

2. Causal effects operate through concentration-dependent rather than elasticity-dependent mechanisms;

3.  The data exhibit exponential growth or decay with respect to predictor levels.

For applications where elasticity interpretations are paramount (e.g., demand curves, production functions), the log-log specification may be more appropriate. However, for domains like gene regulation, chemical kinetics, or systems with threshold effects, our semi-log model provides a more natural representation of the underlying causal mechanisms.

\section{Proof of Proposition 1}

\textbf{Proof.}
By the law of total variance:
\begin{equation}
\text{Var}(X_j) = \mathbb{E}[\text{Var}(X_j | S)] + \text{Var}(\mathbb{E}[X_j | S])
\end{equation}

Since $\mathbb{E}[X_j^2] = \text{Var}(X_j) + (\mathbb{E}[X_j])^2$, and by 
the law of total variance $\text{Var}(X_j) = \mathbb{E}[\text{Var}(X_j | S)] + \text{Var}(\mathbb{E}[X_j | S])$:
\begin{align}
\mathbb{E}[X_j^2] &= \mathbb{E}[\text{Var}(X_j | S)] + \text{Var}(\mathbb{E}[X_j | S]) + (\mathbb{E}[X_j])^2
\end{align}

Using the variance decomposition $\text{Var}(Y) = \mathbb{E}[Y^2] - (\mathbb{E}[Y])^2$ 
with $Y = \mathbb{E}[X_j | S]$:
\begin{equation}
\begin{aligned}
\operatorname{Var}(\mathbb{E}[X_j \mid S]) 
&= \mathbb{E}\!\left[(\mathbb{E}[X_j \mid S])^2\right] 
   - \big(\mathbb{E}[\mathbb{E}[X_j \mid S]]\big)^2 \\
&= \mathbb{E}\!\left[(\mathbb{E}[X_j \mid S])^2\right]
   - (\mathbb{E}[X_j])^2.
\end{aligned}
\end{equation}

where the last equality uses the tower property $\mathbb{E}[\mathbb{E}[X_j | S]] = \mathbb{E}[X_j]$.

Substituting back:
\begin{align}
\mathbb{E}[X_j^2] 
&= \mathbb{E}[\text{Var}(X_j | S)] + \mathbb{E}[(\mathbb{E}[X_j | S])^2]
\end{align}

Therefore:
\begin{equation}
\mathcal{M}(j, S) = \frac{\mathbb{E}[X_j^2]}{\mathbb{E}[(\mathbb{E}[X_j | S])^2]} 
= 1 + \frac{\mathbb{E}[\text{Var}(X_j | S)]}{\mathbb{E}[(\mathbb{E}[X_j | S])^2]}
\end{equation}

(i) Since $\mathbb{E}[\text{Var}(X_j | S)] \geq 0$, we have $\mathcal{M}(j, S) \geq 1$.

(ii) Let $\mathcal{F}_1 = \sigma(S_1)$ and $\mathcal{F}_2 = \sigma(S_2)$ denote the $\sigma$-algebras 
generated by $S_1$ and $S_2$ respectively. Since $S_1 \subseteq S_2$, we have $\mathcal{F}_1 \subseteq \mathcal{F}_2$.

By the definition of conditional variance:
\begin{equation}
\text{Var}(X_j \mid \mathcal{F}) = \mathbb{E}[X_j^2 \mid \mathcal{F}] - (\mathbb{E}[X_j \mid \mathcal{F}])^2
\end{equation}

Taking expectations on both sides:
\begin{align}
\mathbb{E}[\text{Var}(X_j \mid \mathcal{F})] &= \mathbb{E}[\mathbb{E}[X_j^2 \mid \mathcal{F}]] - \mathbb{E}[(\mathbb{E}[X_j \mid \mathcal{F}])^2] \nonumber \\
&= \mathbb{E}[X_j^2] - \mathbb{E}[(\mathbb{E}[X_j \mid \mathcal{F}])^2]
\label{eq:expected_cond_var}
\end{align}
where the second equality uses the tower property $\mathbb{E}[\mathbb{E}[X_j^2 \mid \mathcal{F}]] = \mathbb{E}[X_j^2]$.

Applying \eqref{eq:expected_cond_var} to both $\mathcal{F}_1$ and $\mathcal{F}_2$:
\begin{equation}
\mathbb{E}[\text{Var}(X_j \mid \mathcal{F}_k)] = \mathbb{E}[X_j^2] - \mathbb{E}[(\mathbb{E}[X_j \mid \mathcal{F}_k])^2], \quad k = 1, 2
\end{equation}

Therefore:
\begin{align}
&\mathbb{E}[\text{Var}(X_j \mid \mathcal{F}_1)] - \mathbb{E}[\text{Var}(X_j \mid \mathcal{F}_2)] \nonumber \\
&= \big(\mathbb{E}[X_j^2] - \mathbb{E}[(\mathbb{E}[X_j \mid \mathcal{F}_1])^2]\big) 
   - \big(\mathbb{E}[X_j^2] - \mathbb{E}[(\mathbb{E}[X_j \mid \mathcal{F}_2])^2]\big) \nonumber \\
&= \mathbb{E}[(\mathbb{E}[X_j \mid \mathcal{F}_2])^2] - \mathbb{E}[(\mathbb{E}[X_j \mid \mathcal{F}_1])^2]
\label{eq:var_diff_prelim}
\end{align}

Define $Z := \mathbb{E}[X_j \mid \mathcal{F}_2]$. By Theorem 4.1.13(ii) in \citet{durrett2019probability}, 
since $\mathcal{F}_1 \subseteq \mathcal{F}_2$:
\begin{equation}
\mathbb{E}[Z \mid \mathcal{F}_1] = \mathbb{E}[\mathbb{E}[X_j \mid \mathcal{F}_2] \mid \mathcal{F}_1] = \mathbb{E}[X_j \mid \mathcal{F}_1]
\end{equation}

By Exercise 4.4.5 in \citet{durrett2019probability}:
\begin{equation}
\mathbb{E}[(Z - \mathbb{E}[Z \mid \mathcal{F}_1])^2] = \mathbb{E}[Z^2] - \mathbb{E}[(\mathbb{E}[Z \mid \mathcal{F}_1])^2]
\label{eq:durrett_exercise}
\end{equation}

Substituting $Z = \mathbb{E}[X_j \mid \mathcal{F}_2]$ and $\mathbb{E}[Z \mid \mathcal{F}_1] = \mathbb{E}[X_j \mid \mathcal{F}_1]$:
\begin{align}
&\mathbb{E}[(\mathbb{E}[X_j \mid \mathcal{F}_2] - \mathbb{E}[X_j \mid \mathcal{F}_1])^2] \nonumber \\
&= \mathbb{E}[(\mathbb{E}[X_j \mid \mathcal{F}_2])^2] - \mathbb{E}[(\mathbb{E}[X_j \mid \mathcal{F}_1])^2]
\label{eq:key_identity}
\end{align}

Comparing \eqref{eq:var_diff_prelim} and \eqref{eq:key_identity}, we obtain:
\begin{align}
&\mathbb{E}[\text{Var}(X_j \mid \mathcal{F}_1)] - \mathbb{E}[\text{Var}(X_j \mid \mathcal{F}_2)] \nonumber \\
&= \mathbb{E}[(\mathbb{E}[X_j \mid \mathcal{F}_2] - \mathbb{E}[X_j \mid \mathcal{F}_1])^2]
\label{eq:var_diff_with_square}
\end{align}

On the other hand, by the definition of conditional variance:
\begin{equation}
\text{Var}(Z \mid \mathcal{F}_1) = \mathbb{E}[(Z - \mathbb{E}[Z \mid \mathcal{F}_1])^2 \mid \mathcal{F}_1]
\end{equation}

Taking expectations on both sides:
\begin{equation}
\mathbb{E}[\text{Var}(Z \mid \mathcal{F}_1)] = \mathbb{E}[(Z - \mathbb{E}[Z \mid \mathcal{F}_1])^2]
\label{eq:exp_cond_var_Z}
\end{equation}

Applying \eqref{eq:exp_cond_var_Z} with $Z = \mathbb{E}[X_j \mid \mathcal{F}_2]$ and using $\mathbb{E}[Z \mid \mathcal{F}_1] = \mathbb{E}[X_j \mid \mathcal{F}_1]$:
\begin{equation}
\mathbb{E}[(\mathbb{E}[X_j \mid \mathcal{F}_2] - \mathbb{E}[X_j \mid \mathcal{F}_1])^2]
= \mathbb{E}[\text{Var}(\mathbb{E}[X_j \mid \mathcal{F}_2] \mid \mathcal{F}_1)]
\label{eq:exp_var_of_cond_mean}
\end{equation}

Combining \eqref{eq:var_diff_with_square} and \eqref{eq:exp_var_of_cond_mean}:
\begin{align}
&\mathbb{E}[\text{Var}(X_j \mid \mathcal{F}_1)] - \mathbb{E}[\text{Var}(X_j \mid \mathcal{F}_2)] \nonumber \\
&= \mathbb{E}[\text{Var}(\mathbb{E}[X_j \mid \mathcal{F}_2] \mid \mathcal{F}_1)]
\end{align}

Rearranging:
\begin{equation}
\mathbb{E}[\text{Var}(X_j \mid \mathcal{F}_1)] = \mathbb{E}[\text{Var}(X_j \mid \mathcal{F}_2)] 
+ \mathbb{E}[\text{Var}(\mathbb{E}[X_j \mid \mathcal{F}_2] \mid \mathcal{F}_1)]
\label{eq:total_var_nested}
\end{equation}

Returning to our original notation with $S_1$ and $S_2$:
\begin{align}
&\mathbb{E}[\text{Var}(X_j \mid S_1)] \nonumber \\
&= \mathbb{E}[\text{Var}(X_j \mid S_2)] + \mathbb{E}[\text{Var}(\mathbb{E}[X_j \mid S_2] \mid S_1)] \nonumber \\
&\geq \mathbb{E}[\text{Var}(X_j \mid S_2)]
\end{align}
where the inequality follows since variance is non-negative.

Moreover, by Theorem 4.1.13(ii) in \citet{durrett2019probability}, 
$\mathbb{E}[X_j \mid S_1] = \mathbb{E}[\mathbb{E}[X_j \mid S_2] \mid S_1]$. 
Applying the conditional Jensen inequality to the convex function $\varphi(x) = x^2$:
\begin{align}
&\mathbb{E}[(\mathbb{E}[X_j \mid S_1])^2] \nonumber \\
&= \mathbb{E}[(\mathbb{E}[\mathbb{E}[X_j \mid S_2] \mid S_1])^2] \nonumber \\
&\leq \mathbb{E}[\mathbb{E}[(\mathbb{E}[X_j \mid S_2])^2 \mid S_1]] \nonumber \\
&= \mathbb{E}[(\mathbb{E}[X_j \mid S_2])^2]
\end{align}
where the inequality follows from the conditional Jensen inequality 
$(\mathbb{E}[Y \mid \mathcal{F}])^2 \leq \mathbb{E}[Y^2 \mid \mathcal{F}]$ 
applied to $Y = \mathbb{E}[X_j \mid S_2]$ and $\mathcal{F} = \sigma(S_1)$, 
and the last equality uses the tower property.

Therefore, from the representation in (i), $\mathcal{M}(j, S) = 1 + \frac{\mathbb{E}[\text{Var}(X_j \mid S)]}{\mathbb{E}[(\mathbb{E}[X_j \mid S])^2]}$:
\begin{align}
\mathcal{M}(j, S_2) &= 1 + \frac{\mathbb{E}[\text{Var}(X_j \mid S_2)]}{\mathbb{E}[(\mathbb{E}[X_j \mid S_2])^2]} \nonumber \\
&\leq 1 + \frac{\mathbb{E}[\text{Var}(X_j \mid S_1)]}{\mathbb{E}[(\mathbb{E}[X_j \mid S_1])^2]} \nonumber \\
&= \mathcal{M}(j, S_1)
\end{align}

Thus $\mathcal{M}(j, S_2) \leq \mathcal{M}(j, S_1)$.

(iii) By the local Markov property of DAGs, 
$X_j \perp\!\!\!\perp (\text{NonDesc}(j) \setminus \text{Pa}(j)) \mid \text{Pa}(j)$.

Therefore, for any $S$ satisfying $\text{Pa}(j) \subseteq S$ and 
$S \cap \text{Desc}(j) = \emptyset$ (i.e., $S$ contains all parents but 
no descendants of $j$), we have:
\begin{equation}
\mathbb{E}[X_j | S] = \mathbb{E}[X_j | \text{Pa}(j)]
\end{equation}

This follows because $S \setminus \text{Pa}(j) \subseteq \text{NonDesc}(j) \setminus \text{Pa}(j)$ 
by the condition $S \cap \text{Desc}(j) = \emptyset$, 
which are conditionally independent of $X_j$ given $\text{Pa}(j)$.

Consequently:
\begin{equation}
\text{Var}(X_j | S) = \text{Var}(X_j | \text{Pa}(j)) > 0
\end{equation}
where the strict inequality holds due to the independent noise $\epsilon_j$ 
in model \eqref{eq:log_linear_model}.

This shows $\mathcal{M}(j, S) = \mathcal{M}(j, \text{Pa}(j))$ for all $S$ 
satisfying $\text{Pa}(j) \subseteq S \subseteq \text{NonDesc}(j)$. Combined 
with (ii), this establishes that the moment ratio achieves its minimum value 
on this plateau. $\square$

Now I prove the part 2: Strict inequality when parents are missing

Now consider $S$ such that $\text{Pa}(j) \not\subseteq S$. 
Let $M = \text{Pa}(j) \setminus S \neq \emptyset$ denote the set of missing parents.

\textbf{Key Lemma:} Under the log-linear model, if 
$M \neq \emptyset$ and all $\beta_{kj} \neq 0$ for $k \in \text{Pa}(j)$, then:
\begin{equation}
\mathbb{E}[\text{Var}(X_j | S)] > \mathbb{E}[\text{Var}(X_j | \text{Pa}(j))]
\label{eq:key_inequality}
\end{equation}

\textbf{Proof of Lemma:}

By the conditional variance decomposition for nested $\sigma$-algebras, applied to $\mathcal{F}_1=\sigma(S)$ and $\mathcal{F}_2=\sigma(S \cup \operatorname{Pa}(j))$, we have

$$
\mathbb{E}\left[\operatorname{Var}\left(X_j \mid S\right)\right]=\mathbb{E}\left[\operatorname{Var}\left(X_j \mid S, \operatorname{Pa}(j)\right)\right]+\mathbb{E}\left[\operatorname{Var}\left(\mathbb{E}\left[X_j \mid S, \operatorname{Pa}(j)\right] \mid S\right)\right] .
$$

Since $S \subseteq \operatorname{NonDesc}(j)$, by the local Markov property

$$X_j \perp\!\!\!\perp (S \setminus \operatorname{Pa}(j)) \mid \operatorname{Pa}(j)
$$

which implies
$\mathbb{E}\left[X_j \mid S, \operatorname{Pa}(j)\right]=\mathbb{E}\left[X_j \mid \operatorname{Pa}(j)\right]$ and
$\operatorname{Var}\left(X_j \mid S, \operatorname{Pa}(j)\right)=\operatorname{Var}\left(X_j \mid \operatorname{Pa}(j)\right)$.
Hence

$$
\mathbb{E}\left[\operatorname{Var}\left(X_j \mid S\right)\right]=\mathbb{E}\left[\operatorname{Var}\left(X_j \mid \operatorname{Pa}(j)\right)\right]+\mathbb{E}\left[\operatorname{Var}\left(\mathbb{E}\left[X_j \mid \operatorname{Pa}(j)\right] \mid S\right)\right] .
$$

We need to show that the second term is strictly positive when $M \neq \emptyset$.

From the log-linear model:
\begin{align}
\mathbb{E}[X_j | \text{Pa}(j)] 
&= \mathbb{E}\left[\exp\left(\theta_j + \sum_{k \in \text{Pa}(j)} \beta_{kj} X_k + \epsilon_j\right) \Bigg| \text{Pa}(j)\right] \\
&= \exp\left(\theta_j + \sum_{k \in \text{Pa}(j)} \beta_{kj} X_k\right) \cdot \mathbb{E}[\exp(\epsilon_j)]
\label{eq:cond_mean_parents}
\end{align}

Since $\epsilon_j$ is independent of all parent variables and has non-degenerate 
distribution (with $\text{Var}(\epsilon_j) = \sigma_j^2 > 0$), $\mathbb{E}[\exp(\epsilon_j)]$ 
is a positive constant, say $c_j = \mathbb{E}[\exp(\epsilon_j)] > 0$.

Therefore:
\begin{equation}
\mathbb{E}[X_j | \text{Pa}(j)] = c_j \exp\left(\theta_j + \sum_{k \in \text{Pa}(j)} \beta_{kj} X_k\right)
\label{eq:cond_mean_explicit}
\end{equation}

Now, consider $\text{Var}(\mathbb{E}[X_j | \text{Pa}(j)] | S)$. 

\textbf{Case 1: $S$ contains some but not all parents}

Without loss of generality, partition $\text{Pa}(j) = S_{\text{in}} \cup M$ 
where $S_{\text{in}} = S \cap \text{Pa}(j)$ and $M = \text{Pa}(j) \setminus S \neq \emptyset$.

Then:
\begin{equation}
\mathbb{E}[X_j | \text{Pa}(j)] = c_j \exp\left(\theta_j + \sum_{k \in S_{\text{in}}} \beta_{kj} X_k + \sum_{k \in M} \beta_{kj} X_k\right)
\end{equation}

Conditioning on $S$:
\begin{align}
\mathbb{E}[\mathbb{E}[X_j | \text{Pa}(j)] | S] 
&= c_j \exp\left(\theta_j + \sum_{k \in S_{\text{in}}} \beta_{kj} X_k\right) 
   \cdot \mathbb{E}\left[\exp\left(\sum_{k \in M} \beta_{kj} X_k\right) \Bigg| S\right]
\label{eq:nested_expectation}
\end{align}

The key observation is that $\mathbb{E}[X_j | \text{Pa}(j)]$ depends on the 
missing parents $X_k$ for $k \in M$ through the exponential function. 
Since the exponential is strictly convex and the $X_k$ have non-degenerate 
conditional distributions given $S$ (they are not fixed constants), 
by Jensen's inequality:

\begin{equation}
\text{Var}(\mathbb{E}[X_j | \text{Pa}(j)] | S) > 0
\label{eq:strict_positive_var}
\end{equation}

\textbf{Rigorous justification:}

For any $k \in M$, since $k \in \text{Pa}(j)$ and $\beta_{kj} \neq 0$ by assumption, 
the variable $X_k$ appears in the structural equation for $X_j$ with non-zero 
coefficient. 

Define $Z := \sum_{k \in M} \beta_{kj} X_k$. We need to show that 
$\text{Var}(Z | S) > 0$.

If $\text{Var}(Z | S) = 0$, then $Z$ would be almost surely constant given $S$. 
But this would imply that all $X_k$ for $k \in M$ are deterministic functions 
of $S$ (since the $\beta_{kj} \neq 0$). This contradicts the fact that 
$M \subseteq \text{Pa}(j)$ are true causal parents with their own noise terms 
$\epsilon_k$ that are independent of $S$ by the causal structure.

More formally: Each $X_k$ for $k \in M$ has the form:
\begin{equation}
X_k = \exp\left(\theta_k + \sum_{\ell \in \text{Pa}(k)} \beta_{\ell k} X_\ell + \epsilon_k\right)
\end{equation}

The noise $\epsilon_k$ is independent of all variables not in $\text{Desc}(k)$. 
Since $j \in \text{Desc}(k)$ (as $k \in \text{Pa}(j)$) but $S$ may not fully 
contain $\text{Desc}(k)$, the conditional distribution of $X_k$ given $S$ 
retains variability from $\epsilon_k$.

Therefore, $\text{Var}(Z | S) > 0$, which implies:
\begin{align}
\text{Var}(\mathbb{E}[X_j | \text{Pa}(j)] | S) 
&\geq \text{Var}\left(c_j \exp\left(\theta_j + \sum_{k \in S_{\text{in}}} \beta_{kj} X_k\right) \exp(Z) \Bigg| S\right) \\
&= \left(c_j \exp\left(\theta_j + \sum_{k \in S_{\text{in}}} \beta_{kj} X_k\right)\right)^2 \cdot \text{Var}(\exp(Z) | S) \\
&> 0
\end{align}

where the strict inequality follows because $\text{Var}(\exp(Z) | S) > 0$ when 
$\text{Var}(Z | S) > 0$ (exponential is a non-constant monotone function).

\textbf{Case 2: $S$ contains no parents ($S \cap \text{Pa}(j) = \emptyset$)}

The argument is similar but simpler. The entire term 
$\sum_{k \in \text{Pa}(j)} \beta_{kj} X_k$ is random given $S$, and by the 
same logic, this induces strictly positive conditional variance in 
$\mathbb{E}[X_j | \text{Pa}(j)]$ when conditioned on $S$.

\textbf{Combining the pieces:}

\begin{equation}
\mathbb{E}[\text{Var}(X_j | S)] = \mathbb{E}[\text{Var}(X_j | \text{Pa}(j))] 
+ \mathbb{E}[\text{Var}(\mathbb{E}[X_j | \text{Pa}(j)] | S)]
\end{equation}

We've shown that when $M \neq \emptyset$:
\begin{equation}
\mathbb{E}[\text{Var}(\mathbb{E}[X_j | \text{Pa}(j)] | S)] > 0
\end{equation}

Therefore:
\begin{equation}
\mathbb{E}[\text{Var}(X_j | S)] > \mathbb{E}[\text{Var}(X_j | \text{Pa}(j))]
\end{equation}

Moreover, from part (ii), we have:
\begin{equation}
\mathbb{E}[(\mathbb{E}[X_j | S])^2] \leq \mathbb{E}[(\mathbb{E}[X_j | \text{Pa}(j)])^2]
\end{equation}

Thus,
\begin{align}
\mathcal{M}(j, S) 
&= 1 + \frac{\mathbb{E}[\text{Var}(X_j | S)]}{\mathbb{E}[(\mathbb{E}[X_j | S])^2]} \\
&> 1 + \frac{\mathbb{E}[\text{Var}(X_j | \text{Pa}(j))]}{\mathbb{E}[(\mathbb{E}[X_j | \text{Pa}(j)])^2]} \\
&= \mathcal{M}(j, \text{Pa}(j))
\end{align}

where the strict inequality follows from:
\begin{equation}
\frac{\mathbb{E}[\text{Var}(X_j | S)]}{\mathbb{E}[(\mathbb{E}[X_j | S])^2]} 
> \frac{\mathbb{E}[\text{Var}(X_j | \text{Pa}(j))]}{\mathbb{E}[(\mathbb{E}[X_j | \text{Pa}(j)])^2]}
\end{equation}

because the numerator is strictly larger and the denominator is weakly smaller.

This completes the proof. $\square$

\section{Proof of Proposition 2}

Lemma (Ridge Consistency)

Under Assumption 2(B1)-(B2) and the log-linear model \eqref{eq:log_linear_model}, 
the Ridge estimator $(\hat\theta_j,\hat\beta_j)$ in~\eqref{eq:ridge_objective}
with penalty $\lambda_{\mathrm{ridge}} = o(1)$ as $n \to \infty$ satisfies
$\|\hat\beta_j - \beta_j^*\|_2 = O_p(n^{-1/2})$ and
$|\hat\theta_j - \theta_j^*| = O_p(n^{-1/2})$.

\textbf{Proof.} This follows from standard arguments for regularized 
least-squares regression on the log-scale (see e.g., \cite{buhlmann2011statistics}). 
The sub-Gaussian design (B2) ensures bounded effective rank, while 
sub-exponential responses (B1) provide moment bounds. The rate $O_p(n^{-1/2})$ 
is standard for Ridge with vanishing penalty. $\square$

\textbf{Proposition 2} (Concentration of empirical moment ratios).
Consider the log-linear structural equation model \eqref{eq:log_linear_model}
and fix a node $j$ and candidate set $S$.
Let $\hat{\mathcal{M}}(j,S)$ be the empirical moment-ratio estimator defined in
\eqref{eq:moment_ratio_empirical}, and $\mathcal{M}(j,S)$ its population counterpart
in \eqref{eq:moment_ratio_theoretical}. 
Suppose Assumption 2 holds and the denominator
$\mathbb{E}[(\mathbb{E}[X_j \mid S])^2]$ is bounded away from zero.
Then there exist constants $C_{j,S},c>0$ such that, for all $\delta \in (0,1)$
and all sufficiently large $n$, 
\begin{equation}
\mathbb{P}\Big(
\big|\hat{\mathcal{M}}(j,S) - \mathcal{M}(j,S)\big|
\leq C_{j,S} \sqrt{\tfrac{\log(1/\delta)}{n}}
\Big) \;\geq\; 1 - \delta.
\end{equation}

\begin{proof}
We decompose the proof into five main steps.

\textbf{Step 1: Notation and Decomposition.}

Define:
\begin{itemize}
\item $A := \mathbb{E}[X_j^2]$ and $\hat{A}_n := \frac{1}{n}\sum_{i=1}^n (X_j^{(i)})^2$
\item $\mu_{j|S}^{(i)} := \mathbb{E}[X_j \mid X_S^{(i)}]$ (the true conditional mean at sample $i$)
\item $\hat{\mu}_{j|S}^{(i)} := \exp(\hat{\theta}_j + \sum_{k \in S} \hat{\beta}_{kj} X_k^{(i)})$ (the Ridge prediction)
\item $B := \mathbb{E}[(\mu_{j|S})^2]$ and $\hat{B}_n := \frac{1}{n}\sum_{i=1}^n (\hat{\mu}_{j|S}^{(i)})^2$
\end{itemize}

Then:
\begin{equation}
\hat{\mathcal{M}}(j,S) = \frac{\hat{A}_n}{\hat{B}_n}, \qquad \mathcal{M}(j,S) = \frac{A}{B}
\end{equation}

We have:
\begin{align}
\hat{\mathcal{M}}(j,S) - \mathcal{M}(j,S) 
&= \frac{\hat{A}_n}{\hat{B}_n} - \frac{A}{B} \nonumber \\
&= \frac{B\hat{A}_n - A\hat{B}_n}{B\hat{B}_n} \nonumber \\
&= \frac{B(\hat{A}_n - A) - A(\hat{B}_n - B)}{B\hat{B}_n}
\label{eq:ratio_decomp_proof}
\end{align}

\textbf{Step 2: Concentration of the Numerator $\hat{A}_n$.}

By Assumption 2(B1), $X_j$ is sub-exponential with parameters $(\nu_j, b_j)$. 
This means that for all $t \geq 0$:
\begin{equation}
\mathbb{E}[\exp(t|X_j - \mathbb{E}[X_j]|)] \leq \exp(\nu_j^2 t^2/2) \quad 
\text{for all } t \in [0, 1/b_j]
\end{equation}

Since $X_j$ is sub-exponential, $X_j^2$ is also sub-exponential. Specifically, 
if $Y = X_j^2$, then $Y$ is sub-exponential with parameters $(2\nu_j^2, 2b_j)$.

Applying Bernstein's inequality for sub-exponential random variables 
\citep{vershynin2018high}, we have:
\begin{equation}
\mathbb{P}\left(|\hat{A}_n - A| \geq t\right) \leq 2\exp\left(-c_1 n 
\min\left(\frac{t^2}{\nu_j^4}, \frac{t}{b_j}\right)\right)
\end{equation}

Setting the right-hand side equal to $\delta/3$ and solving for $t$:
\begin{equation}
t = C_1 \sqrt{\frac{\log(1/\delta)}{n}}
\end{equation}
where $C_1$ depends on $(\nu_j, b_j)$.

Therefore, with probability at least $1 - \delta/3$:
\begin{equation}
|\hat{A}_n - A| \leq C_1 \sqrt{\frac{\log(1/\delta)}{n}}
\label{eq:A_concentration_proof}
\end{equation}

\textbf{Step 3: Concentration of the Denominator $\hat{B}_n$.}

We decompose:
\begin{align}
\hat{B}_n - B &= \frac{1}{n}\sum_{i=1}^n (\hat{\mu}_{j|S}^{(i)})^2 - 
\mathbb{E}[(\mu_{j|S})^2] \nonumber \\
&= \frac{1}{n}\sum_{i=1}^n \left[(\hat{\mu}_{j|S}^{(i)})^2 - (\mu_{j|S}^{(i)})^2\right] 
+ \frac{1}{n}\sum_{i=1}^n (\mu_{j|S}^{(i)})^2 - B \nonumber \\
&=: T_1 + T_2
\label{eq:B_decomp_proof}
\end{align}

\textbf{Step 3a: Bounding $T_2$.}

By Assumption 2(B1), $\mu_{j|S}$ is sub-exponential with parameters 
$(\nu_{j,S}, b_{j,S})$. By the same argument as for $\hat{A}_n$, 
$(\mu_{j|S})^2$ is sub-exponential, and applying Bernstein's inequality:
\begin{equation}
|T_2| \leq C_2 \sqrt{\frac{\log(1/\delta)}{n}} \quad 
\text{with probability } 1 - \delta/6
\label{eq:T2_bound_proof}
\end{equation}

\textbf{Step 3b: Bounding $T_1$ (Prediction Error Term).}

From the log-linear model:
\begin{equation}
\mu_{j|S}^{(i)} = \exp\left(\theta_j^* + \sum_{k \in S} \beta_{kj}^* X_k^{(i)}\right) 
\cdot \mathbb{E}[\exp(\epsilon_j)]
\end{equation}
where $(\theta_j^*, \boldsymbol{\beta}_j^*)$ are the population Ridge parameters.

Define the prediction error:
\begin{equation}
e_i := \hat{\mu}_{j|S}^{(i)} - \mu_{j|S}^{(i)}
\end{equation}

We can write:
\begin{align}
\hat{\mu}_{j|S}^{(i)} &= \exp\left(\hat{\theta}_j + \sum_{k \in S} 
\hat{\beta}_{kj} X_k^{(i)}\right) \nonumber \\
&= \exp\left(\theta_j^* + \sum_{k \in S} \beta_{kj}^* X_k^{(i)}\right) 
\cdot \exp\left((\hat{\theta}_j - \theta_j^*) + \sum_{k \in S} 
(\hat{\beta}_{kj} - \beta_{kj}^*) X_k^{(i)}\right)
\end{align}

Using the mean value theorem for the exponential function, for some $\xi$ 
between the true and estimated log-scale predictions:
\begin{equation}
e_i = \mu_{j|S}^{(i)} \cdot \exp(\xi) \cdot \left[(\hat{\theta}_j - \theta_j^*) 
+ \sum_{k \in S} (\hat{\beta}_{kj} - \beta_{kj}^*) X_k^{(i)}\right]
\end{equation}

By Assumption 2(B2), the regressors $X_S$ are sub-Gaussian, so 
$\|X_S^{(i)}\|_2 = O_p(\sqrt{|S|})$. By Cauchy-Schwarz:
\begin{equation}
\left|\sum_{k \in S} (\hat{\beta}_{kj} - \beta_{kj}^*) X_k^{(i)}\right| 
\leq \|\hat{\boldsymbol{\beta}}_j - \boldsymbol{\beta}_j^*\|_2 \cdot \|X_S^{(i)}\|_2
\end{equation}

By Lemma:
\begin{equation}
\|\hat{\boldsymbol{\beta}}_j - \boldsymbol{\beta}_j^*\|_2 = O_p(n^{-1/2}), 
\quad |\hat{\theta}_j - \theta_j^*| = O_p(n^{-1/2})
\end{equation}

Since $\mu_{j|S}^{(i)}$ is sub-exponential (B1), it has bounded exponential 
moments. Specifically, there exists $M_{j,S} > 0$ such that with high probability:
\begin{equation}
\mu_{j|S}^{(i)} \leq M_{j,S}
\end{equation}

Combining these, for each $i$:
\begin{equation}
|e_i| \leq M_{j,S} \cdot \exp(|\xi|) \cdot O_p(n^{-1/2}) \cdot O_p(\sqrt{|S|}) 
= O_p(n^{-1/2})
\end{equation}
uniformly over $i$ with high probability.

Now we bound $T_1$:
\begin{align}
T_1 &= \frac{1}{n}\sum_{i=1}^n \left[(\hat{\mu}_{j|S}^{(i)})^2 - 
(\mu_{j|S}^{(i)})^2\right] \nonumber \\
&= \frac{1}{n}\sum_{i=1}^n (\hat{\mu}_{j|S}^{(i)} + \mu_{j|S}^{(i)})
(\hat{\mu}_{j|S}^{(i)} - \mu_{j|S}^{(i)}) \nonumber \\
&= \frac{1}{n}\sum_{i=1}^n (\hat{\mu}_{j|S}^{(i)} + \mu_{j|S}^{(i)}) \cdot e_i
\end{align}

Using Cauchy-Schwarz:
\begin{align}
|T_1| &\leq \frac{1}{n}\sum_{i=1}^n |\hat{\mu}_{j|S}^{(i)} + \mu_{j|S}^{(i)}| 
\cdot |e_i| \nonumber \\
&\leq \left(\frac{1}{n}\sum_{i=1}^n (\hat{\mu}_{j|S}^{(i)} + \mu_{j|S}^{(i)})^2
\right)^{1/2} \left(\frac{1}{n}\sum_{i=1}^n e_i^2\right)^{1/2}
\end{align}

The first term is $O_p(1)$ (bounded empirical second moment by sub-exponential 
assumption), and the second term is $O_p(n^{-1/2})$ (since $e_i = O_p(n^{-1/2})$ 
uniformly). Therefore:
\begin{equation}
|T_1| = O_p(n^{-1/2})
\end{equation}

This is of smaller order than $\sqrt{\log(1/\delta)/n}$ for fixed $\delta$, 
so for sufficiently large $n$:
\begin{equation}
|T_1| \leq C_3 \sqrt{\frac{\log(1/\delta)}{n}} \quad 
\text{with probability } 1 - \delta/6
\label{eq:T1_final_proof}
\end{equation}

\textbf{Step 3c: Combining for $\hat{B}_n$.}

From \eqref{eq:B_decomp_proof}, \eqref{eq:T2_bound_proof}, and 
\eqref{eq:T1_final_proof}:
\begin{equation}
|\hat{B}_n - B| \leq |T_1| + |T_2| \leq (C_2 + C_3) \sqrt{\frac{\log(1/\delta)}{n}} 
\quad \text{with probability } 1 - \delta/3
\label{eq:B_concentration_proof}
\end{equation}

\textbf{Step 4: Lower Bound on $\hat{B}_n$.}

By assumption, $B = \mathbb{E}[(\mu_{j|S})^2]$ is bounded away from zero: 
$B \geq b_{\min} > 0$.

From \eqref{eq:B_concentration_proof}, with probability $1 - \delta/3$:
\begin{equation}
\hat{B}_n \geq B - (C_2 + C_3)\sqrt{\frac{\log(1/\delta)}{n}} \geq \frac{B}{2} 
\geq \frac{b_{\min}}{2}
\end{equation}
for sufficiently large $n$ (specifically, 
$n \geq \frac{4(C_2+C_3)^2\log(1/\delta)}{B^2}$).

\textbf{Step 5: Final Bound on the Ratio.}

From \eqref{eq:ratio_decomp_proof}:
\begin{align}
|\hat{\mathcal{M}}(j,S) - \mathcal{M}(j,S)| 
&= \left|\frac{B(\hat{A}_n - A) - A(\hat{B}_n - B)}{B\hat{B}_n}\right| \nonumber \\
&\leq \frac{|B(\hat{A}_n - A)| + |A(\hat{B}_n - B)|}{|B\hat{B}_n|} \nonumber \\
&\leq \frac{B \cdot C_1\sqrt{\frac{\log(1/\delta)}{n}} + A \cdot 
(C_2+C_3)\sqrt{\frac{\log(1/\delta)}{n}}}{B \cdot \frac{B}{2}} \nonumber \\
&= \frac{2C_1}{B}\sqrt{\frac{\log(1/\delta)}{n}} + 
\frac{2A(C_2+C_3)}{B^2}\sqrt{\frac{\log(1/\delta)}{n}} \nonumber \\
&\leq C_{j,S} \sqrt{\frac{\log(1/\delta)}{n}}
\end{align}
where $C_{j,S} := \frac{2C_1}{B} + \frac{2A(C_2+C_3)}{B^2}$ depends on the 
sub-exponential parameters and the moments of $X_j$ and $\mu_{j|S}$.

By the union bound over the three probability events (Steps 2, 3a, 3b), 
this holds with probability at least:
\begin{equation}
1 - \left(\frac{\delta}{3} + \frac{\delta}{6} + \frac{\delta}{6}\right) 
= 1 - \frac{2\delta}{3}
\end{equation}

Adjusting the constants appropriately, we can ensure the total failure 
probability is at most $\delta$, completing the proof.
\end{proof}

\section{Proof of Proposition 3}

H-MRS has two computational phases:

\textit{Phase 1 (Ordering):} At each of $p$ ordering steps $m$, we evaluate up to $p - m + 1$ remaining candidates. For each candidate $j$, we fit Ridge regression with at most $m-1$ predictors to compute moment ratios. The total number of Ridge fits is:
\begin{equation}
\sum_{m=1}^p (p - m + 1) \leq p^2
\end{equation}
Each Ridge fit requires $O(nq^2 + q^3)$ operations, yielding $O(p^2 \cdot T_{\text{Ridge}})$ for Phase 1.

\textit{Phase 2 (Parent Selection):} After ordering is complete, we fit ElasticNet once per variable to select parents from predecessors in the ordering. This requires $p$ ElasticNet fits, each with cost $O(nq^2 \cdot K)$ using coordinate descent, yielding $O(p \cdot T_{\text{ElasticNet}})$ for Phase 2.

The total complexity is $O(p^2 \cdot T_{\text{Ridge}} + p \cdot T_{\text{ElasticNet}})$. Since ElasticNet typically requires more iterations than Ridge's closed-form solution, the ElasticNet term may dominate in practice, but the Ridge term scales quadratically with $p$. $\square$

\section{Hyperparameter Selection Guidelines}

This section provides detailed guidance on selecting hyperparameters for H-MRS,
including both theoretical requirements for consistency and practical strategies
for finite-sample settings.

The H-MRS algorithm involves several hyperparameters whose selection affects both 
practical performance and theoretical consistency. We now discuss the role of each 
parameter and provide guidance for their setting.

\textbf{Ridge Regularization Parameter ($\lambda_{\text{ridge}}$).}
The Ridge parameter controls bias-variance tradeoff when estimating 
conditional expectations. For consistency of moment-ratio ordering:
\begin{itemize}
\item \textbf{Theoretical requirement}: $\lambda_{\text{ridge}} \to 0$ and 
$\lambda_{\text{ridge}} \cdot \sqrt{n} \to \infty$ as $n \to \infty$ ensures that 
Ridge estimates converge to true conditional expectations while maintaining stability.
\item \textbf{Practical guidance}: Cross-validation on log-scale prediction error. 
In our experiments, $\lambda_{\text{ridge}} \in [0.01, 1.0]$ performed well.
\item \textbf{Impact on identifiability}: Since moment-ratio minimization (Proposition 1) 
depends only on the ranking of $\mathcal{M}(j,S)$ values, moderate Ridge bias does not 
affect ordering consistency as long as the ranking is preserved.
\end{itemize}

\textbf{ElasticNet Parameters ($\lambda, \rho$).}
These control sparsity in parent selection:
\begin{itemize}
\item \textbf{Theoretical requirement}: For consistency, $\lambda = \lambda_n$ should 
satisfy $\lambda_n \to 0$ and $\lambda_n \sqrt{n} \to \infty$, ensuring both convergence 
to true parameters and correct variable selection under standard ElasticNet theory 
\citep{zou2005regularization}.
\item \textbf{Practical guidance}: 
The mixing parameter $\rho \in [0.5, 0.9]$ balances grouping of correlated parents 
(low $\rho$) and strict sparsity (high $\rho$).
\item \textbf{Impact on identifiability}: The plateau property (Proposition 1(iii)) 
guarantees that all supersets $S \supseteq \text{Pa}(j)$ achieve the same moment ratio. 
ElasticNet's role is to select the minimal such set, which is a model selection problem 
separate from the identifiability established by moment ratios.
\end{itemize}

\textbf{Thresholding Parameter ($\tau$) and Maximum Degree ($d_{\max}$).}
These enforce sparsity in the final graph:
\begin{itemize}
\item \textbf{Theoretical requirement}: $\tau$ should exceed the ElasticNet estimation 
error: $\tau > C \cdot \sqrt{\frac{\log p}{n}}$ for some constant $C$ ensures correct 
parent selection with high probability under standard sparsity assumptions.
\item \textbf{Practical guidance}: Set $\tau$ as a small fraction of median non-zero 
ElasticNet coefficients (e.g., $\tau = 0.1 \cdot \text{median}(|\hat{\beta}_k| : 
\hat{\beta}_k \neq 0)$). The degree constraint $d_{\max}$ should reflect prior knowledge 
about graph sparsity.
\item \textbf{Impact on identifiability}: These parameters do not affect the theoretical 
identifiability of causal ordering (Proposition 1) but control finite-sample edge recovery.
\end{itemize}

\textbf{Summary of Consistency Conditions.}
For asymptotic consistency of H-MRS, hyperparameters should satisfy:
\begin{equation}
\lambda_{\text{ridge}}, \lambda_n \to 0, \quad 
\lambda_{\text{ridge}} \sqrt{n}, \lambda_n \sqrt{n} \to \infty, \quad
\tau \gtrsim \sqrt{\frac{\log p}{n}}
\end{equation}
These conditions ensure: (i) Ridge and ElasticNet estimators converge to population 
parameters, (ii) moment ratios concentrate around their population values (Proposition 2), 
and (iii) sparsity selection is consistent.

\section{Estimated DAG for Financial Data Application}

\begin{figure}[t]
\centering
\includegraphics[width=0.9\columnwidth]{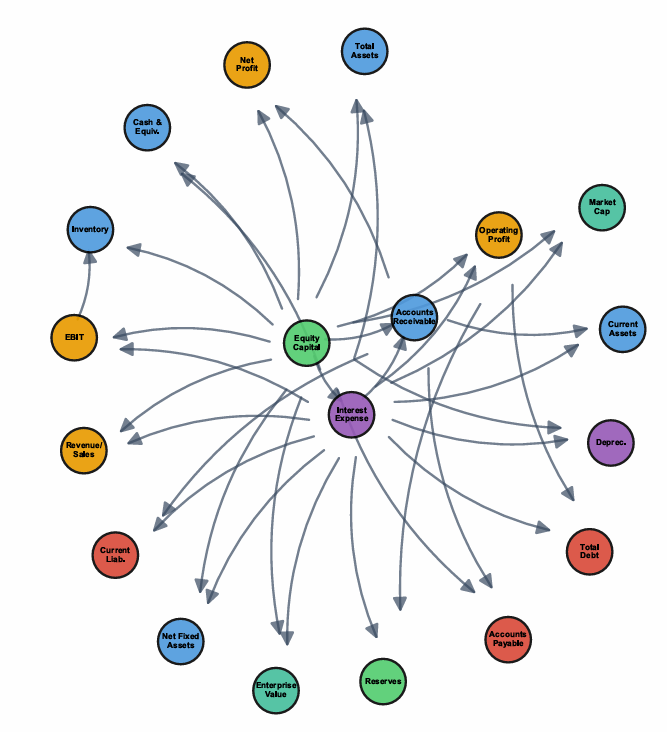}
\caption{Causal structure discovered by H-MRS on financial data (n=2,223 companies, p=19 variables, 35 edges). The algorithm identified Equity Capital and Interest Expense as highly influential nodes with 13 and 15 outgoing edges respectively. Node colors indicate variable categories: blue=Assets, red=Liabilities, green=Equity, orange=Income, purple=Expenses, teal=Valuation.}
\label{fig:financial_dag}
\end{figure}

\end{document}